\documentclass{article}
\usepackage{arxiv}
\usepackage{algorithm}
\usepackage{algorithmic}
\usepackage{amsmath}
\usepackage{amssymb}
\usepackage{array}
\usepackage{multirow}
\usepackage{booktabs}
\usepackage{multirow}
\usepackage{subfigure}
\usepackage{parskip}
\usepackage{cite}
\usepackage[utf8]{inputenc} 
\usepackage[T1]{fontenc}    
\usepackage{hyperref}       
\usepackage{url}            
\usepackage{booktabs}       
\usepackage{amsfonts}       
\usepackage{nicefrac}       
\usepackage{microtype}      
\usepackage{multicol}
\usepackage{lipsum}
\usepackage{graphicx}
\graphicspath{ {./images/} }

\title{PointManifoldCut: Point-wise Augmentation in the Manifold for Point Clouds}

\author{
 Tianfang Zhu \\
  Britton Chance Center for Biomedical Photonics\\
  Huazhong University of Science and Technology \\
  Wuhan, Hubei 430074 \\
  \texttt{ztfun@hust.edu.cn} \\
   \And
 Yue Guan \\
  Britton Chance Center for Biomedical Photonics\\
  Huazhong University of Science and Technology \\
  Wuhan, Hubei 430074 \\
  \texttt{yguan@hust.edu.cn} \\
  \And
 Anan Li \\
  Britton Chance Center for Biomedical Photonics\\
  Huazhong University of Science and Technology \\
  Wuhan, Hubei 430074 \\
  \texttt{aali@hust.edu.cn} \\
}

\begin{document}
\maketitle
\begin{abstract}
Mixed-based point cloud augmentation is a popular solution to the problem of limited availability of large-scale public datasets.
But the mismatch between mixed points and corresponding semantic labels hinders the further application in point-wise tasks such as part segmentation.
This paper proposes a point cloud augmentation approach, PointManifoldCut(PMC), 
which replaces the neural network embedded points, rather than the Euclidean space coordinates.
This approach takes the advantage that points at the higher levels of the neural network are already trained to embed its neighbors relations
and mixing these representation will not mingle the relation between itself and its label.
We set up a spatial transform module after PointManifoldCut operation to align the new instances in the embedded space.
The effects of different hidden layers and methods of replacing points are also discussed in this paper.
The experiments show that our proposed approach can enhance the performance of point cloud classification as well as segmentation networks, and brings them additional robustness to attacks and geometric transformations.
The code of this paper is available at: 
https://github.com/fun0515/PointManifoldCut.
\end{abstract}

\begin{multicols}{2}
\section{Introduction}
\label{sec:intro}
Rapidly developing 3D reconstruction technologies such as LiDAR scanners
and other sensors~\cite{lidar} can directly generate massive point clouds,
which are widely used in fields,
including autonomous driving~\cite{auto-drive},
geographical mapping~\cite{geography} and so on.
To meet the needs of these applications, 
a series of data-driven approaches for learning features 
from point clouds have been proposed in recent years~\cite{review1,review2}.
These methods can be used for several visual tasks, 
such as
point cloud classification~\cite{cls1,pointnet,pointnet2,dgcnn},
point cloud segmentation~\cite{seg1,seg2,seg3}
as well as 
point cloud generation~\cite{gen1,gen2}.
However, over-fitting and poor generalization~\cite{review2} caused by 
the dwarf scale of the existing public point cloud datasets~\cite{modelnet,shapenet} 
is one of the challenges faced by many point cloud processing and analysis methods.

Inspired by the mixed sample data augmentation in image domain~\cite{mixup,manifoldmixup,cutmix}, 
researchers have been exploring augmentation methods to create new training data 
from existing point clouds in recent years.
PointMixup~\cite{PointMixup} defines a data augmentation between the point clouds 
as a shortest path linear interpolation, which solves the problem of point mismatch, and can be applied to the object-level point cloud classification.
PointCutMix~\cite{PointCutMix} directly replaces the point sets in two point clouds to obtain new training data. Since no new points are generated in this process, it can also be used for point cloud segmentation in addition to object-level classification.
However, the change of neighborhood may cause the distortion of point-wise label 
and hinder the improvement on the network performance, 
due to the semantic information of points not only comes from themselves, 
but also from their neighborhood in the point cloud. 

In this paper, we note that many learning-based point cloud processing methods constantly 
update the features of each point from the neighbor nodes~\cite{dgcnn,pointnet2},
so the node features in a high-dimensional manifold contain the knowledge about itself as well as its neighborhood, 
which are more compatible with the semantic labels than the original Euclidean spatial coordinates.
From this observation, this paper proposes a point-wise data augmentation method, named PointManifoldCut,
which replaces the points with the corresponding labels from different samples on manifold to generate new training data.
This technique is verified in point cloud classification and segmentation tasks respectively,
and it achieves the cutting-edge results in both tasks.
In order to align new samples generated in the embedded space, a spatial transform module always follows the PointManifoldCut operation.
The replacement strategy of points and the insertion location of PointManifoldCut are discussed in this paper.
We also show that this data augmentation
method is insensitive to the point drop, coordinates noise and other geometric transformations,
and it is attractive to be applied to a broader range of point
cloud learning tasks.

\makeatletter
\newenvironment{figurehere} 
    {\def\@captype{figure}} 
    {} 
    
\newenvironment{tablehere}
    {\def\@captype{table}}
    {}
\makeatother

\begin{figurehere}
\centering
\includegraphics[width=1.0\columnwidth]{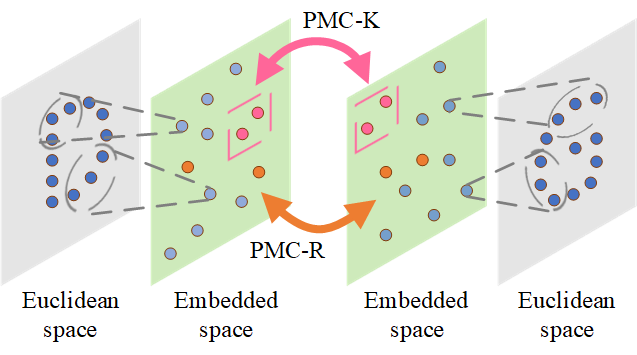}
\caption{\textbf{PointManifoldCut: replace points in the manifold.}
The \textbf{gray} and \textbf{green} represent the point clouds in Euclidean space and their corresponding manifolds 
in the embedded space 
respectively.
Points in the embedded space contain their neighborhood information owe to the learning process of deep networks.
PMC-K and PMC-R are two replacement methods described in Sec.~\ref{sec:alg}.
}
\label{fig:cut}
\end{figurehere}

\section{Related work}
\label{sec:related work}
\textbf{Deep learning for point clouds.}
As a flexible geometric representation, 
there is no ordering in the point cloud data, 
thus it is permutation invariant,
which prevents 2D convolutional neural networks(CNN) from 
being applied directly on point clouds.
PointNet~\cite{pointnet} is the pioneering work 
that uses deep learning network to learn from the unordered points,
where the global feature is learned by the shared multi-layer perceptions 
and a max pooling operator.
After that, PointNet++~\cite{pointnet2} improves the performance 
by introducing the neighbor information missing from PointNet.
It learns the geometric features by sampling 
and partition the local regions at multi-level scale.
Similarly, recent work has focused on capturing local features 
in an efficient approach~\cite{shellnet,pointcnn,pointconv,sonet,densepoint,rgcnn}.
For instance, DGCNN~\cite{dgcnn} dynamically builds multi-layer local graphs 
and updates node semantic characteristics in the feature space.
Besides, 
\cite{mvcnn} uses two-dimensional image sets to describe point clouds, 
and then captures the features like in the image domain.
In general, the high-dimensional features for one point learned 
by the multi-layer neural network contain richer semantic information 
than the original Euclidean coordinates of the specific point.

\textbf{Mixed-based data augmentation.}
Enhancing model performance by data augmentation is common practice 
in real-world application due to its effectiveness.
Mixup~\cite{mixup} uses the weighted interpolation of two images 
to generate new training data,
while CutMix~\cite{cutmix} directly cuts and swaps the corresponding regions of the two images,
where the weight is determined by the size of the clipping region.
These works are the two example ideas for image augmentation,
and some work has extended them on the hidden respresentations 
such as Manifold Mixup~\cite{manifoldmixup} and PatchUp~\cite{patchup}.
Several empirical studies show 
that these data amplification techniques are able to improve 
the performance of deep neural networks for 2D vision task.

\textbf{Data augmentation for point clouds.}
Because the common augmentation techniques such as rotation 
and jittering ignore the shape complexity of point clouds,
some mixed-based point cloud augmentation methods are proposed over the past few years.
In order to solve the problem that points do not correspond one by one in point clouds,
PointMixup~\cite{PointMixup} defines data augmentation for point clouds as the shortest path linear interpolation,
which can be seen as the implementation of Mixup~\cite{mixup} on point clouds.
However, this interpolation will generate unlabeled points, 
which makes it is hard to apply it to point cloud segmentation.
Inspired by CutMix~\cite{cutmix}, 
PointCutMix~\cite{PointCutMix} directly replace the partial set of points between two samples.
Since the original labels can be migrated along with points, 
this method can be used in the point cloud segmentation task as well.

However, the change of neighbors may cause the distortion of the point-wise semantic label thus hurt the model performance, because the semantic information of a point comes from not only itself but also its surrounding points.
For example, a single point in a digit cannot represent a complete mathematical meaning as shown in Fig.~\ref{fig:cut}.
This paper aims to propose a point-wise point clouds augmentation 
that allows to match the replaced labels.

\begin{figure*}[t]
\centering
\includegraphics[width=2.0\columnwidth]{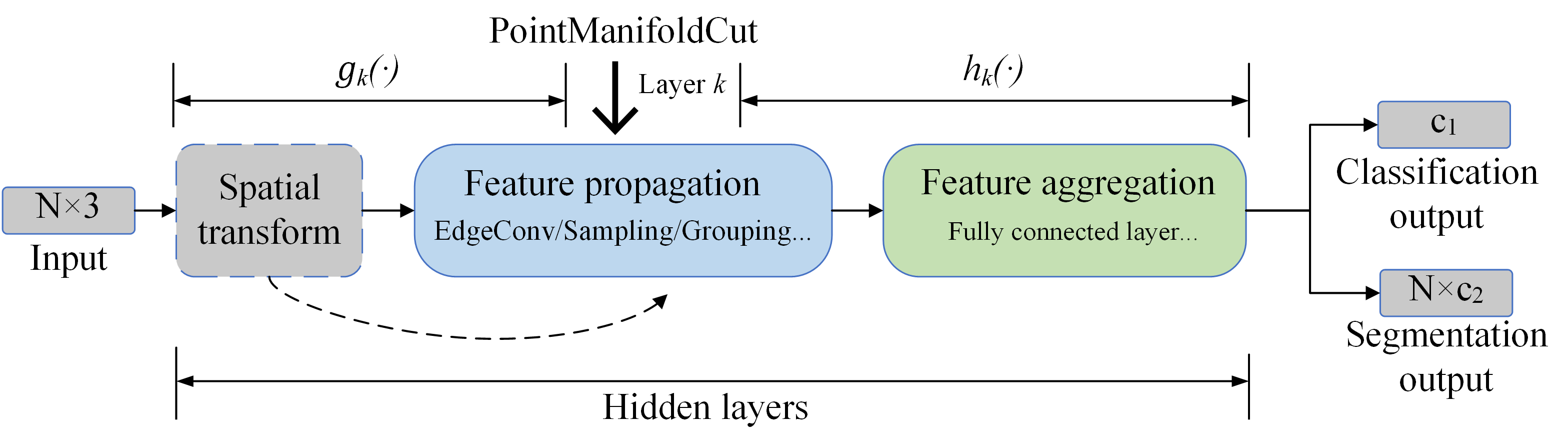} 
\caption{\textbf{A typical point cloud neural network 
for point cloud classification and semantic segmentation.}
It can be divided into the propagation stage, where shape features are captured,
and the aggregation stage that decodes features into the output score vectors.
Some network example components are presented in the figure, 
such as EdgeConv in DGCNN and sampling layers in PointNet++.
Spatial transformation is a frequently employed operator 
to align point clouds in the Euclidean space. 
We recommend using PointManifoldCut in the propagation phase 
and moving spatial transform after PointManifoldCut to align the new samples in the embedded space.}
\label{fig:layer}
\end{figure*}

\section{PointManifoldCut}
\label{sec:pmc}
\subsection{Problem statement}
\label{sec:setting}
Here we focus on two basic point cloud processing tasks: point cloud classification 
and point cloud segmentation, 
while the latter can be regarded as point-wise classification.
A training set $\mathbb{P}$ containing $J$ point clouds is given as $\mathbb{P}=\{(P_j,c_j,S_j)\}^J_{j=1}$, where $P_j=\{p_n^j\}^N_{n=1}\in P$ is a point cloud sample consisting of $N$ points;
$c_j\in\{0,1\}^{C_1}$ is the one-hot class label for a total of $C_1$ classes of point clouds; and $S_j=\{s_n^j\}^N_{n=1}$, where $s_n^j\in\{0,1\}^{C_2}$, is the one-hot class label for a total of $C_2$ classes for each point in one point cloud.
Usually the initial feature of the point is the 3D coordinates of Euclidean space, as $p_n^j\in\mathbb{R}^3$. 

A point cloud classification task aims to train a function $h_c$: $P\rightarrow[0,1]^{C_1}$ that maps a point cloud to a object-level semantic label distribution. So the function $h_c(\cdot)$ can be learned by minimizing the loss function, as shown in \ref{equ:cls}
\begin{equation}
    \theta^{*}_c=\mathop{argmin}\limits_{\theta_c} \sum_{P\in \mathbb{P}}L(h_c(P),c)
\label{equ:cls}
\end{equation}
, where $\theta^{*}_c$ is the optimal parameter and $L$ denotes the loss function such as the cross-entropy loss function used in this work.
Likewise, the goal of a typical point cloud segmentation task is to learn a function $h_s$: $(P,c)\rightarrow[0,1]^{C_2}_N$ that maps a point cloud to a point-wise semantic label distribution.
The optimization for $\theta_s$ is similar as shown in \ref{equ:seg}.
\begin{equation}
    \theta^{*}_s=\mathop{argmin}\limits_{\theta_s} \sum_{P\in \mathbb{P}}L(h_s(P,c),s^N)
\label{equ:seg}
\end{equation}
Here, $h_c(P)$ and $h_s(P,c)$ respectively represent the output of classification networks and segmentation networks.

\subsection{Algorithm}
\label{sec:alg}
PointManifoldCut can be viewed as a data augmentation
applied in the embedded space for the vanilla point cloud
learning networks. 
It creates a new point cloud instance $(g_k(\tilde{P}),\tilde{c},\tilde{S})$ through the substitution of points in manifold domain from two available samples, where $g_k(\cdot)$ is the mapping from the input data to the hidden representation at hidden layer $k$.
So the neural network $h_c(\cdot)$ in \eqref{equ:cls} and $h_s(\cdot)$ in \eqref{equ:seg} can be rewriten as $h(P)=h_k(g_k(P))$, where subscript is dropped for easier reading.
Here, $h_k(\cdot)$ denotes the mapping from the
hidden representation layer $k$ to the output as shown in Fig.~\ref{fig:layer}.

The PointManifoldCut operation, $PMC(b, b', M)$, for batch $b$, shuffled batch $b'$, and a diagonal matrix mask $M$ at hidden layer $k$ can be defined for two available training samples $(P_1,c_1,S_1)$ and $(P_2,c_2,S_2)$ without lossing generality, as shown in \ref{equ:algorithm}:
\begin{equation}
\begin{aligned}
    g_k(\tilde{P})&=M\cdot g_k(P_1)+(I-M)\cdot g_k(P_2) \\
    \tilde{c}&=\lambda c_1+(1-\lambda) c_2\\
    \tilde{S}&=M\cdot S_1 + (I_N-M)\cdot S_2
\label{equ:algorithm}
\end{aligned}
\end{equation}
, where $M=diag\{m_1,m_2,...,m_N\}$ and $m_i$ is a boolean element indicating from which instance the $i^{th}$ point is selected, which is like the mask used in vision task.
$I$ is an identity matrix with $N$ diagonal elements. 
$\lambda \in [0,1]$ and it follows the $Beta(\beta,\beta)$ distribution, indicating the ratio of selected points from the two training samples.

Furthermore, hyperparameter $\rho \in[0, 1]$ is introduced to precisely control the possibility that PointManifoldCut will be used. 
When $\rho=0$, point clouds will not be augmented.
Otherwise PointManifoldCut will be used for all the point clouds when $\rho=1$.
Hence the two kinds of loss function can be mathematically denoted as in \ref{equ:loss}:
\begin{equation}
\begin{aligned}
    L_c(h)&=\sum_{p\in \mathbb{P}}(1-\rho)L(h(P),c)+\rho L(h(\tilde{P}),\tilde{c})\\
    L_s(h)&=\sum_{p\in \mathbb{P}}(1-\rho)L(h(P,c),S)+\rho L_s(h(\tilde{P},\tilde{c}),\tilde{S})
\label{equ:loss}
\end{aligned}
\end{equation}
, where $L_c$ and $L_s$ respectively represent the loss function of point cloud classification and segmentation task.
Algorithm~\ref{alg:pmc} shows the general process of PointManifoldCut during training, where the $Uniform(0,1)$ represents generate a random number that follows the Uniform distribution and $diag(a,b)$ represents a diagonal matrix where the first $a$ elements are 1 and the remaining $b$ elements on the diagonal are 0.
$PMC(\cdot)$ is the procedure described in \ref{equ:algorithm}, while $shuffle(\cdot)$ is a randomly permutated version of its argument.
$1(M)$ is the function to find how many element $1$ are there in the diagonal matrix $M$.

\begin{algorithm}[H]
\caption{PointManifoldCut(PMC)}
\label{alg:pmc}
\textbf{Input}: Deep networks $h$, training batch $b:(i,t)$ where $i$ and $t$ are the input and target of $h$.\\
\textbf{Parameter}: Threshold $\rho \in[0,1]$, $\beta>0$ and the number of points $n$. \\ 
\textbf{Output}: Value $l$ of criterion $L$. \\
\textbf{Process}:
\begin{algorithmic}[1] 
\STATE $\theta=Uniform(0,1)$  
\IF {$\theta<\rho$}
\STATE $\lambda=Beta(\beta,\beta)$
\STATE Initialize $M \leftarrow diag(\lfloor\lambda \times n \rfloor,\lfloor(1-\lambda) \times n \rfloor)$
\STATE New batch $\tilde{b}:(\tilde{i},\tilde{t}) \leftarrow PMC(b,shuffle(b),M)$
\STATE Update $\lambda \leftarrow 1(M)/n$
\STATE $l \leftarrow L(h(\tilde{i})),\tilde{t}))\times \lambda+L(h(\tilde{i}),t)\times (1-\lambda)$
\ELSE
\STATE $l \leftarrow L(h(i),t)$ 
\ENDIF
\STATE \textbf{return} $l$
\end{algorithmic}
\end{algorithm}

\subsection{Analysis}
\label{sec:ana}

\textbf{How to choose the layer $k$}.
Although PointManifoldCut can be used after any hidden layer of point cloud processing networks, we recommend using it in the propagation stage of features(Fig~\ref{fig:layer}).
The augmentation should not be applied too late in networks to retain sufficient non-linear layers from the final output.
However, it is hard to know exactly which hidden layer has enough information to represent a new sample, especially when it depends on various neural networks,
thus some work in 2D vision field~\cite{manifoldmixup,patchup} choose to randomly select a hidden layer and apply augmentation techniques.
We find that this random strategy can achieve an average performance of selecting fixed layers in point cloud networks(see Sec.~\ref{sec:cls} and \ref{sec:seg}).
According to this structure, PointCutMix~\cite{PointCutMix} can be regarded as a special case of our methods at the input layer.

\textbf{Does the replacement method of points matter?}
There are two different replacement methods for point clouds. 
As shown in Fig.~\ref{fig:cut}, one is to randomly select a series of points for replacement(PointManifold-R), the other is to select a random central point in the second cloud and replace its neighbor points into the first point cloud(PointManifold-K).
As the new sample obtained by the second replacement method in Euclidean space conserves the semantic relations among its neighbors, so it has better effect than random replacement~\cite{PointCutMix}.
However, every point in the manifold is far away from other points. 
Therefore, the nearest neighbors replacement loses its advantage in this scenario, and the effect of random replacement is better, which is supported in Sec.~\ref{sec:cls}.

\textbf{T-net after PointManifoldCut.} 
T-net~\cite{tnet} is a mini-network for learning affine transformation matrix and usually takes the original point clouds as input~\cite{pointnet,dgcnn}.
We set the T-net after PointManifoldCut operation instead of immediately after input to unify the transformation invariance of new samples in manifold, which can be described as:
\begin{equation}
    t(g_k(\tilde{p}))=A \times g_k(\tilde{p}), A \in \mathbb{R}^{d \times d}
\label{equ:t-net}
\end{equation}
where $t$ and $A$ respectively denotes the spatial transform operation and the transformation matrix output by T-net. 
$d$ is the feature dimension of layer $k$.
In PointNet, a regularization term is set to restrict the transformation matrix to a form that is close to the orthogonal matrix, thus reducing the difficulty of optimization. 
This is also be used in our work, as described as:
\begin{equation}
    L_{reg}=||I-AA^T||^2
\label{equ:lreg}
\end{equation}
where $L_{reg}$ represent the regularization term added to the training loss and $I$ is an identity matrix of the same size as transformation matrix $A$.

\section{Experiments}
In this section, we compare the performance improvement of PointManifoldCut in classification task as well as segmentation task against the existing point cloud augmentation methods, and show its enhancement of network robustness. 
We also report the effects of two hyperparameters~($\beta$,$\rho$) and the location which PointManifoldCut is applied in the network. 
To ensure universality, all experiments are based on multiple point cloud networks.
\subsection{Experiments Setup}
\label{subsec:expsetup}

\textbf{Datasets.}
We use two point cloud datasets in our experiments.
ModelNet40~\cite{modelnet} contains 12311 CAD samples from 40 object categories, 
while each sample consists of 2048 points.
Among them, 9843 models are used for training and the other 2468 models are left out as the test set.
ShapeNet~\cite{shapenet} parts contains 16880 samples from 16 categories and 50 part labels, it is split into 14006 for training and 2874 for testing.
Both datasets are publicly available, 
and we use the ModelNet40 for the point cloud classification experiments
while run point cloud segmentation experiments on the ShapeNet dataset.

\textbf{Networks.}
To show that PointManifoldCut is a general technique and agnostic to the network
which is employed, we experiment with three cutting edge network architectures, including PointNet~\cite{pointnet}, PointNet++~\cite{pointnet2}, and DGCNN~\cite{dgcnn}, where PointNet learns the features based solely on one point
and does not consider the neighbor information, 
PointNet++ obtains multi-level local features by hierarchical sampling, 
and DGCNN builds dynamic local graphs at each layer based on the k-nearest neighbors(KNN) method.

\textbf{Implementation details.}
We implement our work in PyTorch~\cite{pytorch}.
All the networks used in the experiments take 2048 points as an input point cloud instance.
Each experiment runs for 250 epoches with a batch size of 48 for classification and 16 for segmentation, on a NVIDIA Quadro RTX 5000 GPU card.
We use the Adam~\cite{adam} optimizer with an initial learning rate of 0.001 that decays by half every 20 epoches to train the PointNet and PointNet++.
DGCNN is trained with the SGD~\cite{sgd} optimizer with an initial learning rate of 0.1, 
which is reduced until 0.001 through cosine annealing.
The momentum for batch normalization of SGD is 0.9 and
the batch normalization decay is not used.

\begin{tablehere}
\centering
\caption{\textbf{Overall accuracy(\%) with different augmentations on Modelnet40.} PointManifoldCut-K and PointManifold-R represent the results where the nearest neighbors replacement and random replacement used respectively.}
\begin{tabular}{lp{0.9cm}<{\centering}p{1.2cm}<{\centering}p{1.0cm}<{\centering}}
\toprule
Method      & PointNet & PointNet++ & DGCNN \\ \midrule
Baseline    & 89.2     & 90.7       & 92.3  \\
PointMixup  & 89.9     & 92.7       & 93.1  \\
PointCutMix & 89.9     & \textbf{93.4}       & 93.2  \\  \midrule
PointManifoldCut-K        & 89.4             & 92.9 & 93.1 \\
PointManifoldCut-R        & \textbf{90.0}     & 93.0       & \textbf{93.7}  \\
\bottomrule
\end{tabular}
\label{tab:cls}
\end{tablehere}

\begin{figurehere}
\centering
\includegraphics[width=1.0\columnwidth]{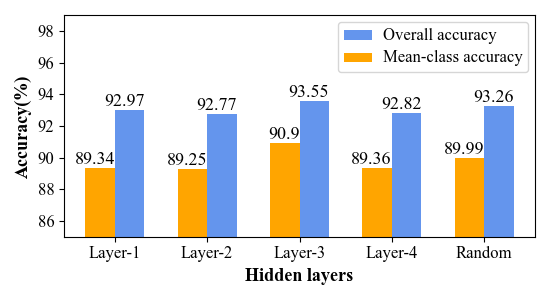} 
\caption{\textbf{Performance comparison 
when PointManifoldCut is applied to layer $k$.}
The last two columns represent the result of selecting a random hidden layer per batch to use PointManifoldCut.
$\rho=1$ and $\beta=0.5$ are used in this experiment.}
\label{fig:clslayer}
\end{figurehere}

\subsection{Classification}
\label{sec:cls}
We perform experiments on ModelNet40 dataset to show how PointManifoldCut 
improves the performance in object-level task 
with respect to 
the different locations in the network where PointManifoldCut is applied to. 

\textbf{Comparison to other augmentation methods.}
Here, we compare the improvements brought by PointManifoldCut 
and other point cloud augmentation technologies, 
including PointMixup~\cite{PointMixup} and PointCutMix~\cite{PointCutMix}, on several classification networks. From Table~\ref{tab:cls}, it can be seen that PointManifoldCut improves the classification performance of DGCNN most significantly, 
about 1.4\%, 0.6\% and 0.5\% higher than PointMixup and PointCutMix.
For PointNet, the three augmentations show similar improvement over the vanilla model.
When the baseline network is PointNet++, 
the effect of PointManiflodCut is between PointMixup and PointCutMix, 
0.3\% higher than the former and 0.4\% lower than the latter.
Besides, we compare the two replacement schemes of points introduced in Sec.~\ref{sec:ana}, 
where PointManifoldCut-K and PointManifoldCut-R represent the nearest neighbors replacement and random replacement used respectively.
It is shown that random replacement is generally better than nearest neighbors replacement in the manifold.
 In general, our PointManfoldCut improves the overall accuracy of cutting-edge point cloud classification networks by 0.8\% to 2.3\%, 
 which is competitive compared with the existing methods.

\begin{table*}[t]
\centering
\caption{\textbf{Improvements of segmentation with different augmentations on ShapeNet part dataset.} The Metric is mIoU(\%) on points.
PCM indicates the PointCutMix augmentation method.
The first row is the sample size of the corresponding category.}
\begin{tabular}{p{1.3cm}|p{0.6cm}<{\centering}|p{0.5cm}<{\centering}p{0.35cm}<{\centering}p{0.35cm}<{\centering}p{0.35cm}<{\centering}p{0.5cm}<{\centering}p{0.5cm}<{\centering}p{0.35cm}<{\centering}p{0.5cm}<{\centering}p{0.5cm}<{\centering}p{0.4cm}<{\centering}p{0.5cm}<{\centering}p{0.35cm}<{\centering}p{0.5cm}<{\centering}p{0.35cm}<{\centering}p{0.5cm}<{\centering}p{0.5cm}<{\centering}} \toprule
Method       & mIoU          & \begin{tabular}[c]{@{}c@{}}Air\\ plane\end{tabular} & Bag           & Cap           & Car           & Chair         & \begin{tabular}[c]{@{}c@{}}Ear\\ phone\end{tabular} & \begin{tabular}[c]{@{}c@{}}Gui-\\tar \end{tabular}         & Knife         & Lamp          & \begin{tabular}[c]{@{}c@{}} Lap-\\top \end{tabular}       & \begin{tabular}[c]{@{}c@{}}Motor\\ bike\end{tabular} & Mug           & Pistol        & \begin{tabular}[c]{@{}c@{}}Roc-\\ket \end{tabular}       & \begin{tabular}[c]{@{}c@{}}Skate\\ board\end{tabular} & Table         \\ \midrule
             &               & 2690                                                 & 76            & 55            & 898           & 3758          & 69                                                   & 787           & 392           & 1547          & 451           & 202                                                   & 184           & 283           & 66            & 152                                                    & 5271          \\ \midrule
PointNet     & 83.1          & 82.2                                                 & 72.3          & 79.1          & 72.0          & 89.2          & 72.9                                                 & 90.0          & 84.8          & 78.2          & 95.0          & 62.3                                                  & 89.9          & 81.2          & 52.4          & 73.1                                                   & 81.9          \\
+PCM & 83.7          & 82.0                                                 & 70.6          & 77.6          & 75.1          & 89.6          & \textbf{73.9}                                        & 90.3          & 85.9          & 78.8          & 95.1          & 63.8                                                  & \textbf{91.8} & 80.3          & 53.7          & 72.7                                                   & 82.4          \\
+Ours        & \textbf{84.2} & \textbf{83.2}                                        & \textbf{76.7} & \textbf{80.7} & \textbf{75.9} & \textbf{90.0} & 72.5                                                 & \textbf{90.2} & \textbf{86.0} & \textbf{79.6} & \textbf{95.3} & \textbf{66.3}                                         & 91.1          & \textbf{82.2} & \textbf{55.5} & \textbf{73.3}                                          & \textbf{82.8} \\ \midrule
PointNet2   & 84.9          & 82.1                                                 & 83.0          & 82.3          & 77.4          & 90.2          & 69.9                                                 & 90.8          & 86.9          & 84.3          & 95.2          & 68.6                                                  & 94.1          & 82.7          & 60.2          & 74.4                                                   & 82.8          \\
+PCM & \textbf{85.5} & 82.6                                                 & \textbf{85.9} & 83.7          & \textbf{78.3} & 90.7          & 72.5                                                 & 90.9          & \textbf{87.7} & 84.3          & \textbf{95.3} & \textbf{70.7}                                         & \textbf{95.1} & 82.4          & \textbf{62.3} & \textbf{74.9}                                          & \textbf{83.4} \\
+Ours        & 85.3          & \textbf{82.6}                                        & 85.2          & \textbf{84.9} & 78.1          & \textbf{90.9} & \textbf{73.3}                                        & \textbf{91.3} & 87.1          & \textbf{84.4} & 95.2          & 69.9                                                  & 94.4          & \textbf{82.9} & 58.6          & 74.5                                                   & 82.6          \\ \midrule
DGCNN        & 85.2          & 84.0                                                 & 83.4          & 86.7          & 77.8          & 90.6          & 74.7                                                 & 91.2          & 87.5          & 82.8          & 95.7          & 66.3                                                  & \textbf{94.9}          & 81.1          & \textbf{63.5}          & 74.5                                                   & 82.6          \\
+PCM & 85.7          & 84.1                                                 & \textbf{84.5} & 87.6          & \textbf{78.1} & 91.0          & \textbf{77.7}                                        & 91.6          & \textbf{88.9} & 84.4          & 95.9          & 68.9                                                  & 94.6 & \textbf{83.1} & 60.3 & \textbf{75.6}                                          & \textbf{83.0} \\
+Ours        & \textbf{85.7} & \textbf{84.2}                                        & 83.3          & \textbf{87.8} & 77.6          & \textbf{91.1} & 73.3                                                 & \textbf{91.6} & 88.7          & \textbf{84.7} & \textbf{96.0} & \textbf{71.5}                                         & 94.4          & 82.4          & 59.8          & 73.8                                                   & 82.9     \\ \bottomrule    
\end{tabular}
\label{tab:seg}
\end{table*}

\textbf{The layer $k$ where PointManifoldCut is applied to.}
We further investigate the impact of site where PointManifoldCut is inserted into the DGCNN
on the model performnance.
DGCNN dynamically constructs local graphs 
and learns features through four consecutive EdgeConv~\cite{dgcnn} layers.
Fig.~\ref{fig:clslayer} shows the overall accuracy and mean-class accuracy of using PointManfoldCut after these four different EdgeConv layers or a random EdgeConv layer.
It can be seen that the effect is similar for each case.
But when PointManfoldCut is applied to latter layer~(layer-3), 
the overall accuracy and mean class accuracy are higher 
than the second performance by 0.6\% and 0.5\% respectively.
In addition, selecting a random layer per batch achieves an average effect, which is practical if it is not clear which layer is the most suitable.

Besides, we compare the two replacement schemes of points introduced in Sec.~\ref{sec:ana}, 
where PointManifoldCut-K and PointManifoldCut-R represent the nearest neighbor replacement and random replacement used respectively.
It is shown that random replacement is generally better than nearest neighbor replacement in the manifold.

\subsection{Part segmentation}
\label{sec:seg}
Point cloud segmentation is another important task in point cloud learning.
We discuss the impact of point-wise features at different levels on PointManifoldCut, and then compare its performance improvement with existing method on ShapeNet parts dataset.
Since PointMixup lacks interpolated point labels, the comparison here focuses on PointCutMix.
We choose mIoU(mean Intersection-over-Union) as the evaluation metric of point cloud segmentation task, which is calculated by averaging the IoUs of all the testing shapes.

\textbf{Comparison to other augmentation method.}
Here, we show in detail the improvement effect of PointManifoldCut on the point-wise classification task of three deep networks, including PointNet, PointNet++ and DGCNN, as well as the comparison with PointCutMix.
From Table~\ref{tab:seg}, a notable improvement of PointManifoldCut on the segmentation metrics of PointNet can be seen, with a total increase of 1.1\%, which is 0.5\% higher than that of PointCutMix.
Since PointNet hardly capture the local information, this promotion does not come from the smooth combination of samples, but because the combination of samples in the manifold domain are more meaningful.
However, PointManifoldCut has limited improvement on PointNet++ and lags behind PointCutMix by 0.3\%. 
It may because that we only replace points after feature inverse interpolation, resulting in insufficient learning of new samples.
For DGCNN, PointManifoldCut and PointCutMix both improve the performance of segmentation by about 0.5\%.

\textbf{The  layer $k$ where  PointManifoldCut  is  applied to.}
We firstly compare the influences of replacing differents level of point-wise features on point cloud segmentation tasks.
DGCNN uses three EdgeConv layers to capture local features in segmenting branch, while PointNet also sets up a three-layer perceptron before concatenating the category vector.
So we use PointManifoldCut in the first three hidden layers to see if there is a significant impact.
Because the sampling layers of PointNet++ continuously reduces the scale of point set, PointManifoldCut for PointNet++ is only used after all points get high-dimensional features through inverse interpolation.

\begin{tablehere}
\centering
\caption{\textbf{mIoU when PointManfoldCut applied on different hidden layers.}
The Follow-T column lists the value when PointManfoldCut is applied after T-net; The Lead-T column shows the value when PointManfoldCut is applied in front of T-net.}
\begin{tabular}{lcccc}
\toprule
\multirow{2}{*}{Layers} & \multicolumn{2}{c}{PointNet} & \multicolumn{2}{c}{DGCNN} \\ \cline{2-5} \rule{0pt}{11pt} 
                  & Follow-T        &   Lead-T      & Fllow-T       & Lead-T      \\ \midrule
Layer-1           & 83.40         & \textbf{84.25}        & 85.52        & \textbf{85.71}      \\
Layer-2           & 83.52         & 83.63        & 85.30        & 85.44      \\
Layer-3           & 83.31         & 83.70        & 85.51        & 85.53     \\ 
Random            & 82.91         & 84.15        & 85.31
 & 85.69      \\ \midrule
Baseline          & \multicolumn{2}{c}{83.13}    & \multicolumn{2}{c}{85.23} \\ \bottomrule
\end{tabular}
\label{tab:kseg}
\end{tablehere}

At the same time, we compare the effect of setting the spatial transform before or after using PointManifoldCut operation.
As shown in Table~\ref{tab:kseg}, different-level hidden representation has greater influence on segmentation task than that of classification task, which may be that segmentation task is more sensitive to point-wise features.
We also observed a similar trend in the improvement effect of PointManifoldCut on PointNet and DGCNN, which is the most significant in layer-1, then decreased in layer-2 and rebounded in layer-3.
This may indicate that the features of the first hidden layer represent the most meaningful new samples constructed.
Similar to Fig.~\ref{fig:layer}, random selection of a hidden layer achieves an average improvement in the point cloud segmentation task.
Besides, setting spatial transform after PointManifoldCut is always better than before.

\subsection{Hyperparameter tuning}
Here we explore the effects of two hyperparameters $\rho$ and $\beta$ on PointManifoldCut.
We take the classification task as an example and the results are shown in Fig.~\ref{fig:hyper}.

\begin{figurehere}
\centering
    \subfigure[]{
        \includegraphics[width=3.2in]{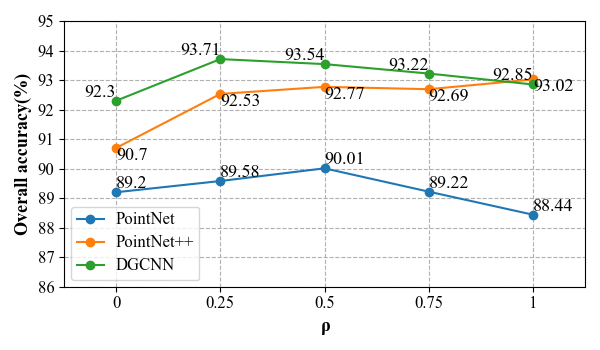}
        \label{fig:alpha}
    }
    \quad    
    \subfigure[]{
	\includegraphics[width=3.2in]{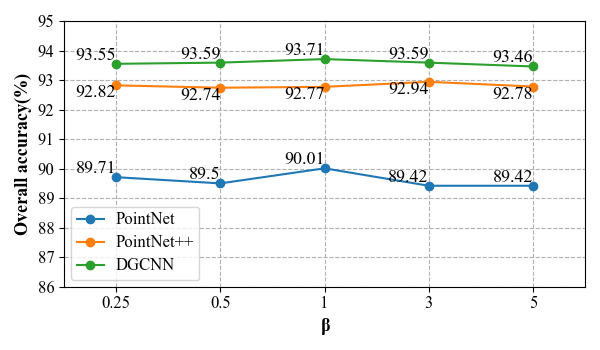}
        \label{fig:beta}
    }
\caption{\textbf{Hyperparameter tuning for $\rho$ and $\beta$ on ModelNet40.}
PointManifoldCut is applied to three classification networks respectively.
$\beta=1$ in the top plot and $\rho=0.5$ in the bottom plot.}
\label{fig:hyper}
\end{figurehere}

\textbf{For $\rho$.} 
Hyperparameter $\rho$ introduced in Sec.~\ref{sec:alg} determines 
how many samples need to be replaced by PointManifoldCut in the cost function.
We test with the $\rho$ values evenly from 0 to 1,
while $\beta$ is fixed to 1 
and plot the classification performance of three point cloud networks in Fig.~\ref{fig:alpha}. 

The classification accuracy of DGCNN and PointNet employing PointManifoldCut 
reach the best performance~(93.71\% and 90.01\%) 
when $\rho$ is set to 0.25 and 0.5 respectively, 
and then decreases as the $\rho$ increases.
When all training samples are augmented($\rho=1$), 
the accuracy for PointNet is even 0.8\% lower than 
that without PointManifold operation~($\rho=0$).
However, PointNet++ experiments shows an opposite trend.
Its overall accuracy increases slowly with the probability of PointManifoldCut being used, 
and finally reaches the highest~(93.02\%) when $\rho$ is set to 1.

\textbf{For $\beta$.} 
As mentioned in Sec.~\ref{sec:alg}, 
$\beta$ controls the number of replaced points in the point clouds.
We fixed $\rho$ to 0.5 and set different $\beta$ values to 
explore its impact on the performance for the point cloud networks.
Fig.~\ref{fig:beta} suggests that performance are not as sensitive to the hyperparameter $\beta$ as to $\rho$.
The classification accuracy of DGCNN and PointNet++ 
remains almost constant in the selected interval of $\beta$.
Although the fluctuation of accuracy for PointNet is larger, it does not exceed 0.5\% either.

\begin{figurehere}
\centering
\subfigure[]{
        \includegraphics[width=1.0\columnwidth]{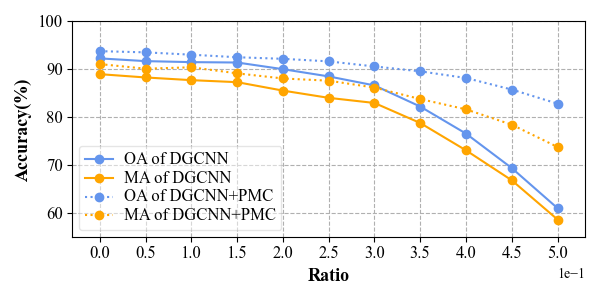}
        \label{fig:line_point_drop}
    }
    \quad    
    \subfigure[]{
	\includegraphics[width=1.0\columnwidth]{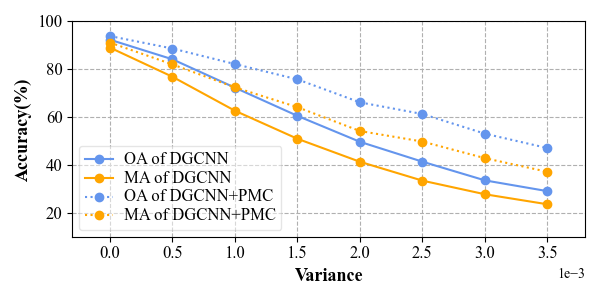}
        \label{fig:line_noise}
    }
\caption{\textbf{Classification performance under two attacks.}
The above is the result of randomly discarding points, and the below is the result of adding noise to the coordinates.}
\label{fig:noisycls}
\end{figurehere}

\subsection{Robustness test}
In addition to the direct improvement of our augmentation method on the performance of deep neural network in point cloud classification and segmentation tasks, we further discuss its impact on the robustness of these networks.
We focus on two common point cloud attacks: point drop and adding random noise, as well as geometric transformations, including scaling and rotation.
Point drop~\cite{pointdropping} refers to the random dropping of a portion of points in a point cloud random, and PointDrop(0.2) means the dropped points account for one fifth of the total points.
Similar to image domain, gaussian noise with different variance is randomly added to the three coordinates of points to attack the networks.
For geometric transformations, Scale(0.2) and X-rotation($30^o$) denote enlarging the point cloud by 1.2 times and rotating it $30^o$ around the X-axis, respectively.
We apply these attacks and transformations on the test set, and then report the performance of the networks trained with the normal training set. 
Below are the results.

\begin{tablehere}
\centering
\caption{\textbf{Robustness test on ModelNet40.}
Var is the variance of the added gaussian noise.
Metrics are overall accuracy(OA) and mean-class accuracy(MA) of classification task.}
\begin{tabular}{p{2.4cm}p{0.7cm}<{\centering}p{0.7cm}<{\centering}p{0.9cm}<{\centering}p{0.9cm}<{\centering}}
\toprule
\multirow{2}{*}{Attack} & \multicolumn{2}{c}{DGCNN} & \multicolumn{2}{c}{\begin{tabular}[c]{@{}c@{}}DGCNN+\\PointManfoldCut\end{tabular}} \\ \cline{2-5} \rule{0pt}{12pt} 
                        & OA(\%)      & MA(\%)      & OA(\%)                                    & MA(\%)                                   \\ \midrule
PointDrop(0.2)          & 89.95       & 85.48       & \textbf{92.11}                                     & \textbf{88.01}                                    \\
PointDrop(0.3)          & 86.56       & 82.90       & \textbf{90.52}                                     &\textbf{86.11}                                    \\
PointDrop(0.4)          & 76.51       & 73.02       & \textbf{88.11}                                     & \textbf{81.57}                                    \\ \midrule
Noise var=0.001         & 72.18       & 62.56       & \textbf{82.03}                                     & \textbf{72.31}                                    \\
Noise var=0.002         & 49.59       & 41.26       & \textbf{66.05}                                     & \textbf{54.09}                                    \\
Noise var=0.003         & 33.50       & 27.71       & \textbf{52.98}                                     & \textbf{42.76}   \\ \midrule
Scale(0.6)              & 72.34       & 63.29       &\textbf{84.52}
                                 & \textbf{78.10}   \\
Scale(0.8)              & 89.67       & 86.09       & \textbf{91.91}
                                 & \textbf{88.59}   \\
Scale(1.4)              & 82.84  & 78.94            & \textbf{86.32}
                                 & \textbf{81.13}   \\ \midrule
X-rotation($30^o$)      & 54.94       & \textbf{53.03}       & \textbf{56.05}
                                 & 51.50   \\
Y-rotation($30^o$)      & \textbf{83.33}       & \textbf{80.38}       & 81.17
                                 & 78.98   \\
Z-rotation($30^o$)      & 51.02  & 44.16            & \textbf{55.31}
                                 & \textbf{46.34}   \\ \bottomrule                        
\end{tabular}
\label{tab:rub1}
\end{tablehere}

\textbf{Point cloud classification.}
We conducted 12 attacks on the test set of ModelNet40 in Table~\ref{tab:rub1}, and then show the classification accuracy of DGCNN with and without PointManifoldCut training.
It can be seen that PointManifold brings remarkable additional robustness to DGCNN on the whole, and it becomes more obvious as the attack intensifies.
This trend is clearly shown in Fig.~\ref{fig:noisycls}.
When one fifth of the points are dropped randomly, PointManifoldCut only improves the overall accuracy of 2.2\% and the mean-class accuracy of 2.5\%. 
But when the dropping ratio increased to 0.4, the two indicators increased by 11.6\% and 8.6\% respectively.
In the face of gaussian noise with variance of 0.001, PointManifoldCut can improve the overall accuracy as well as mean-class accuracy by nearly 10\%.
When the variance increases to 0.003, the overall accuracy of the original DGCNN decreases to 33.5\%, but it can be stabilized at about 53.0\% after using PointManifoldCut, and the increment is about 19.5\%.
However, PointManifoldCut brings no additional robustness to rotation transformation.
For example, if the point clouds rotate $30^o$ around the X-axis in Euclidean space, the overall accuracy and mean-class accuracy of DGCNN will decay to 54.94\% and 53.03\%, respectively.
But PointManifoldCut can only slightly increase the overall accuracy by about 1.1\%, and reduce the mean-class accuracy by about 1.5\% at the same time.
This may be due to the lack of special alignment of new samples in the embedded space for classification networks.

\begin{tablehere}
\centering
\caption{\textbf{Robustness test on ShapeNet part dataset.}
Var is the variance of the added gaussian noise.
The metric is mIoU(\%).}
\begin{tabular}{lp{0.95cm}<{\centering}p{0.9cm}<{\centering}p{0.9cm}<{\centering}p{0.9cm}<{\centering}}
\toprule
\multirow{2}{*}{Attack} & \multicolumn{2}{c}{PointNet} & \multicolumn{2}{c}{DGCNN} \\ \cline{2-5} \rule{0pt}{12pt}
                     & Baseline       & +PMC        & Baseline      & +PMC      \\ \midrule
Noise var=0.001                & 77.55          & \textbf{79.18}       & 76.48         & \textbf{79.78}     \\
Noise var=0.002                & 73.91          & \textbf{75.88}       & 72.11         & \textbf{75.93}     \\
Noise var=0.003                & 70.84          & \textbf{72.85}       & 68.41         & \textbf{72.65}    \\ \midrule
Scale(0.6)  & 82.60    & \textbf{82.75}    & \textbf{83.88}    &83.24 \\
Scale(0.8)  & 83.01    & \textbf{83.86}    & 84.82    &\textbf{85.52} \\
Scale(1.4)  & 83.02    & \textbf{83.61}    & \textbf{84.12}    &83.67 \\ \midrule
X-rotation($30^o$) &\textbf{59.75} &52.62 &62.93 &\textbf{69.94} \\
Y-rotation($30^o$) &70.79 &\textbf{71.36} &78.45 &\textbf{78.63} \\
Z-rotation($30^o$) &48.66 &\textbf{55.70} &62.62 &\textbf{72.33} \\ \bottomrule
\end{tabular}
\label{tab:2robu}
\end{tablehere}

\textbf{Point cloud segmentation.}
Then we evaluate the impact of PointManifoldCut on the robustness of segmentation tasks. 
Similar to the classification task, PointManifoldCut also brings remarkable robustness improvement in segmentation(see Table~\ref{tab:2robu}).
For PointNet, the performance of PointManifoldCut under Gaussian noise is around 2\% higher than that of the original network.
And for DGCNN, PointManifoldCut brings an additional 3.3\% improvement when the variance is 0.001. 
And when the variance increases to 0.003, the mIoU of the baseline is 68.4\%, which increases by 4.3\% after using PointManifoldCut.
In contrast, scaling does less damage to the point cloud segmentation results.
When the size of point cloud is reduced from 0.8 to 0.6, the segmentation metrics of PointNet and DGCNN only decrease by about 0.4\% and 1.0\% respectively.
But if the point clouds rotate $30^o$ around the Z-axis in Euclidean space, the segmentation performance of PointNet and DGCNN will decay to 48.66\% and 62.62\%, respectively.
However, PointManifold can bring about 7\%-9\% improvements, making them stable at 55.70\% and 72.33\%.

Fig.~\ref{fig:noisyseg} shows the visual segmentation results of DGCNN before and after using PointManifoldCut under gaussian noise with different intensity.
It can be clearly seen that although the part segmentation effect of DGCNN on the aircraft model is close to the ground truth without gaussian noise, there are still some bad judgments. 
For example, an end point of the wing is wrongly judged as the fuselage, which is corrected by PointManifoldCut.
In addition, gaussian noise does harm to the shape of point clouds, which puts forward higher requirements for the robustness of deep neural networks.
In the face of gaussian noise with variance of 0.001, the original DGCNN loses part of the wing between the engine and the fuselage, which can be segmented after using PointManifoldCut.
When the variance increases to 0.003, the segmentation ability of DGCNN decreases rapidly.
It identifies all wings and engines as other categories, and almost all tails as engines.
However, after using PointManifoldCut, the fuselage, wing, tail and engine are identified and roughly matched.

\begin{figurehere}
\centering
\includegraphics[width=1.0\columnwidth]{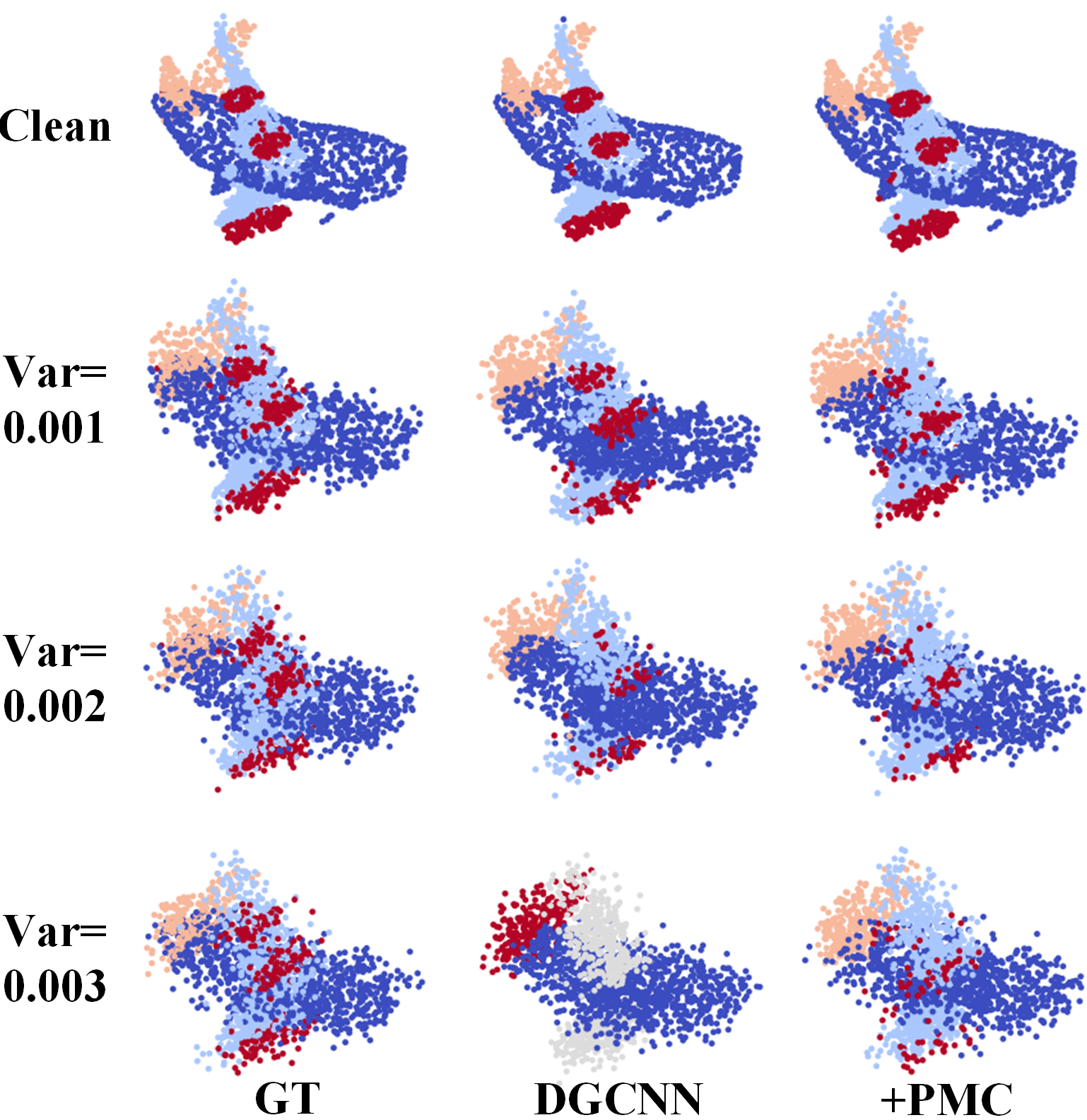} 
\caption{\textbf{Segmentation results of DGCNN on ShapeNet part dataset.}
Each row represents the input with different variance gaussian noise.
The first column is the ground truth, the second column is the output of the original DGCNN, while the third column is the output with PointManifoldCut. 
The color illustrates the category of the part to which the point belongs.}
\label{fig:noisyseg}
\end{figurehere}

\section{Conclusion}
We presented PointManifoldCut, a point-wise augmentation method for point clouds, which explores the points mix-up options and the possible site that the method is applied.
The experiments suggest that our method outperforms other point augmentation methods
on cutting-edge point cloud neural networks not only for point cloud classification task, but also for segmentation task.
However, a future research direction is to embed the structural information of the point clouds
during the data augmentation
to further regularize the point cloud learning models.
\end{multicols}

\bibliographystyle{unsrt}  


\bibliography{qhe}

\end{document}